\documentclass{article}
\usepackage{ijcai22}

\usepackage{times}
\usepackage{soul}
\usepackage{url}
\usepackage[hidelinks]{hyperref}
\usepackage[utf8]{inputenc}
\usepackage[small]{caption}
\usepackage{graphicx}
\usepackage{amsmath}
\usepackage{amsthm}
\usepackage{booktabs}
\usepackage{algorithm}
\usepackage{algorithmic}

\usepackage{epstopdf}
\usepackage{enumitem}

\title{TransPPG: Two-stream Transformer for Remote Heart Rate Estimate}

\author{
	Jiaqi Kang$^1$
	\and
	Su Yang$^1$\and
	Weishan Zhang$^{2}$
	\affiliations
	$^1$Shanghai Key Laboratory of Intelligent Informatoin Processing, School of Computer Science, Fudan University\\
	$^2$School of Computer Science, China University of Petroleum\\
	\emails
	\{jqkang19, suyang\}@fudan.edu.cn,
	zhangws@upc.edu.cn
}

\begin{document}
	
	\maketitle
	
	\begin{abstract}
		Non-contact facial video-based heart rate estimation using remote photoplethysmography (rPPG) has shown great potential in many applications (e.g., remote health care) and achieved creditable results in constrained scenarios. However, practical applications require results to be accurate even under complex environment with head movement and unstable illumination. Therefore, improving the performance of rPPG in complex environment has become a key challenge. In this paper, we propose a novel video embedding method that embeds each facial video sequence into a feature map referred to as Multi-scale Adaptive Spatial and Temporal Map with Overlap (MAST\_Mop), which contains not only vital information but also surrounding information as reference, which acts as the mirror to figure out the homogeneous perturbations imposed on foreground and background simultaneously, such as illumination instability. Correspondingly, we propose a two-stream Transformer model to map the MAST\_Mop into heart rate (HR), where one stream follows the pulse signal in the facial area while the other figures out the perturbation signal from the surrounding region such that the difference of the two channels leads to adaptive noise cancellation. Our approach significantly outperforms all current state-of-the-art methods on two public datasets MAHNOB-HCI and VIPL-HR. As far as we know, it is the first work with Transformer as backbone to capture the temporal dependencies in rPPGs and apply the two stream scheme to figure out the interference from backgrounds as mirror of the corresponding perturbation on foreground signals for noise tolerating.
	\end{abstract}
	
	\section{Introduction}
	
	\noindent Heart rate (HR) is one of the most significant vital signals that reflects people's physical status. HR monitoring plays an important role in a variety of applications such as health care, early detection of cardiovascular diseases, and characterizing emotional status in psychology tests. Traditional heart rate measurement generally utilizes Electrocardiography (ECG) and Photoplethysmograph (PPG) \cite{ksc:pulse}, both of which need professional equipments to attach sensors to people's skin, in general discomfortable or inconvenient. To avoid these problems, remote photoplethysmograph (rPPG) \cite{vsn:remote} is proposed, the target of which is to recover the volumetric change of blood over time through facial videos captured by webcams remotely, based on measuring subtle color changes of the skin. Here, the resulting signal is known as blood volume pulse (BVP), from which HR can be estimated. However, the extracted BVPs will be significantly disturbed under realistic situations, because changes of skin colors in videos are caused by both vital (blood volume) and non-vital factors (e.g., illumination and head movement), and the amplitude of the latter is much greater than the former, leading to very low signal-to-noise ratio. Therefore, extracting BVPs from facial videos has been remaining a challenging task so far.
	\begin{figure}[t]
		\centering
		\includegraphics[width=0.9\columnwidth]{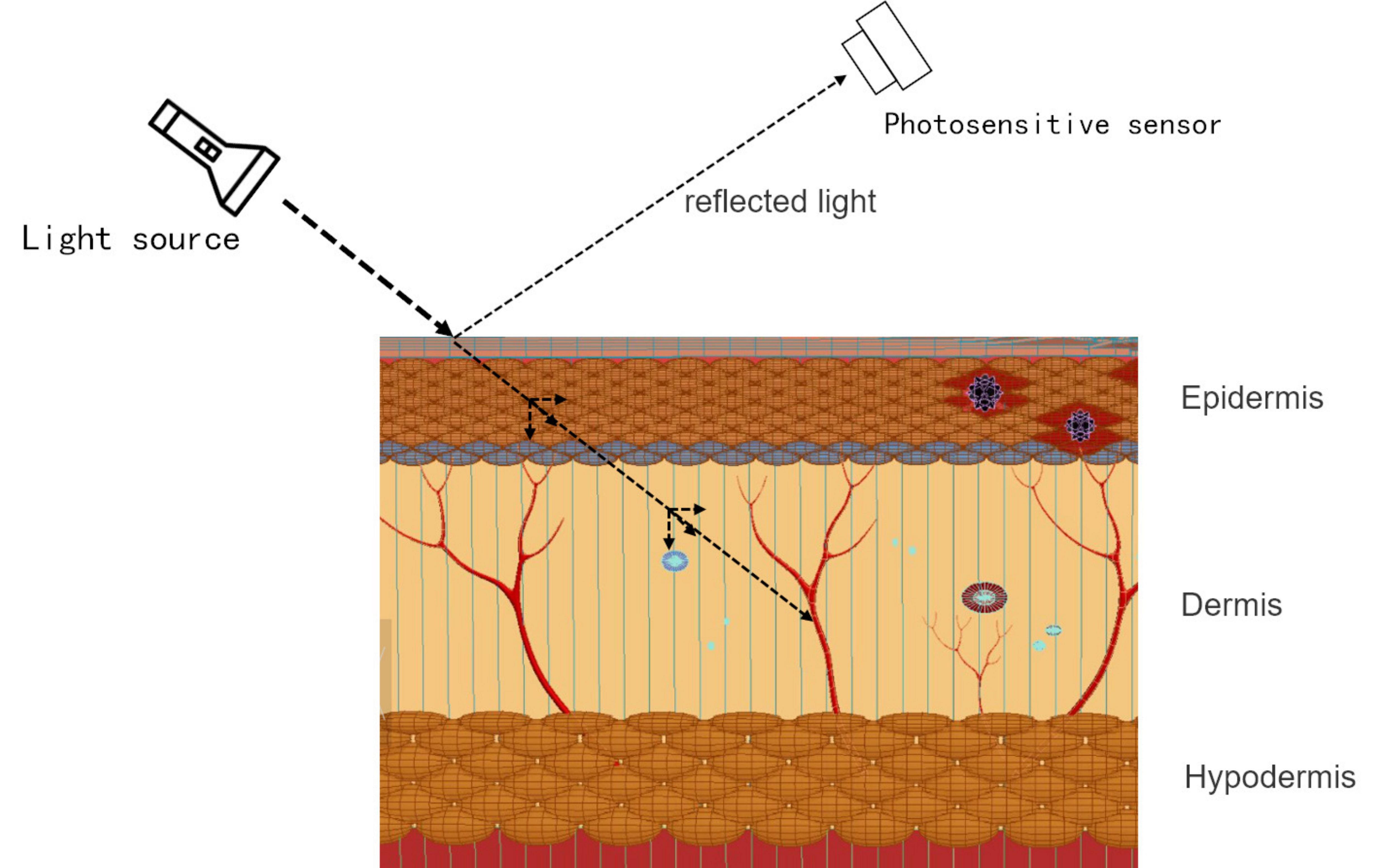}
		\caption{Simplified skin reflection model. When a beam of light irradiates on the surface of the skin, a part of it is absorbed by different tissues while the other part is reflected}.
		\label{fig1}
	\end{figure}
	
	In recent years, many rPPG-based methods have been proposed towards obtaining more accurate and stable HR value. \cite{pmp:no}  utilized independent component analysis (ICA) to obtain cleaner BVP signals. CHROME \cite{dj:robust} and POS \cite{wdsd:pos} transformed the color from RGB space to a new one based on prior knowledge to avoid the adverse impact of illumination conditions. \cite{tar:sr} selected regions of interest (ROIs) adaptively to compute BVP based on matrix completion. Recently, several deep learning based approaches were proposed. \cite{vfm:visual}  first extracted features via spatial decomposition and temporal filtering from particular face region, and then utilized CNN to map the features to HR value. \cite{cm:deepphys} designed a convolutional attention network that takes the normalized difference between frames as input to predict BVP signals. \cite{nshc:rhythmnet} proposed a novel spatial-temporal feature referred to as MSTMap, which is extracted from multiple small face ROIs, and then utilized a CNN-RNN model for HR estimation. \cite{yplhz:stven} proposed a two-stage CNN that first enhance the quality of videos, and then utilized the enhanced videos to obtain more stable BVP. \cite{nyhlsz:cvdnet} utilized disentangle representation learning to denoise the raw BVP signals extracted from videos so as to deal with unstable environmental factors.
	
	However, all the aforementioned approaches have some common defects: First, they are only focused on detecting the pulse patterns from face region (say foreground)  and ignore the information contained in the background, which is the key for noise removal since the signal from background can act as a mirror to figure out the homogeneous perturbation imposed on foreground face region such as illumination caused disturbing. Second, during data processing, they crop and resize the face region to solve the problem of inconsistent face size, and utilize facial landmarks to align face regions in order to solve the head movement problem. However, all the resize and align functions will change pixel values from the original frames, which will introduce new noise. Third, almost all the learning-based approaches utilize CNN to estimate HR, which is not good at handling temporal information. Therefore, in this paper, we propose a new framework to address these problems. Our contributions are as follows:
	\begin{itemize}[fullwidth,itemindent=2em]
		\item[(1)] We propose a novel video embedding method that can embed videos into feature maps referred to as MAST\_Mop, which can solve the problem of inconsistent face size and head movement without introducing any noise since no resize operation is needed for the sake of alignment.
		\item[(2)] We propose a two-stream Transformer network that utilize both foreground information and background information to estimate HR, where the background stream functions to perceive interference arising from background so as to reduce its negative impact on foreground signals. As far as we konw, it is the first work with Transformer as backbone to capture the temporal dependencies in rPPGs and apply the two-stream scheme to figure out the interference from backgrounds as mirror of the corresponding perturbation on foreground signals.
		\item[(3)] The proposed approach has achieved state-of-the-art performances on two public datasets MAHNOB-HCI \cite{slpp:hci} and VIPL-HR \cite{nhsc:vipl}.
	\end{itemize}
	
	\begin{figure*}[t]
		\centering
		\includegraphics[width=0.7\textwidth,height=0.2\textheight]{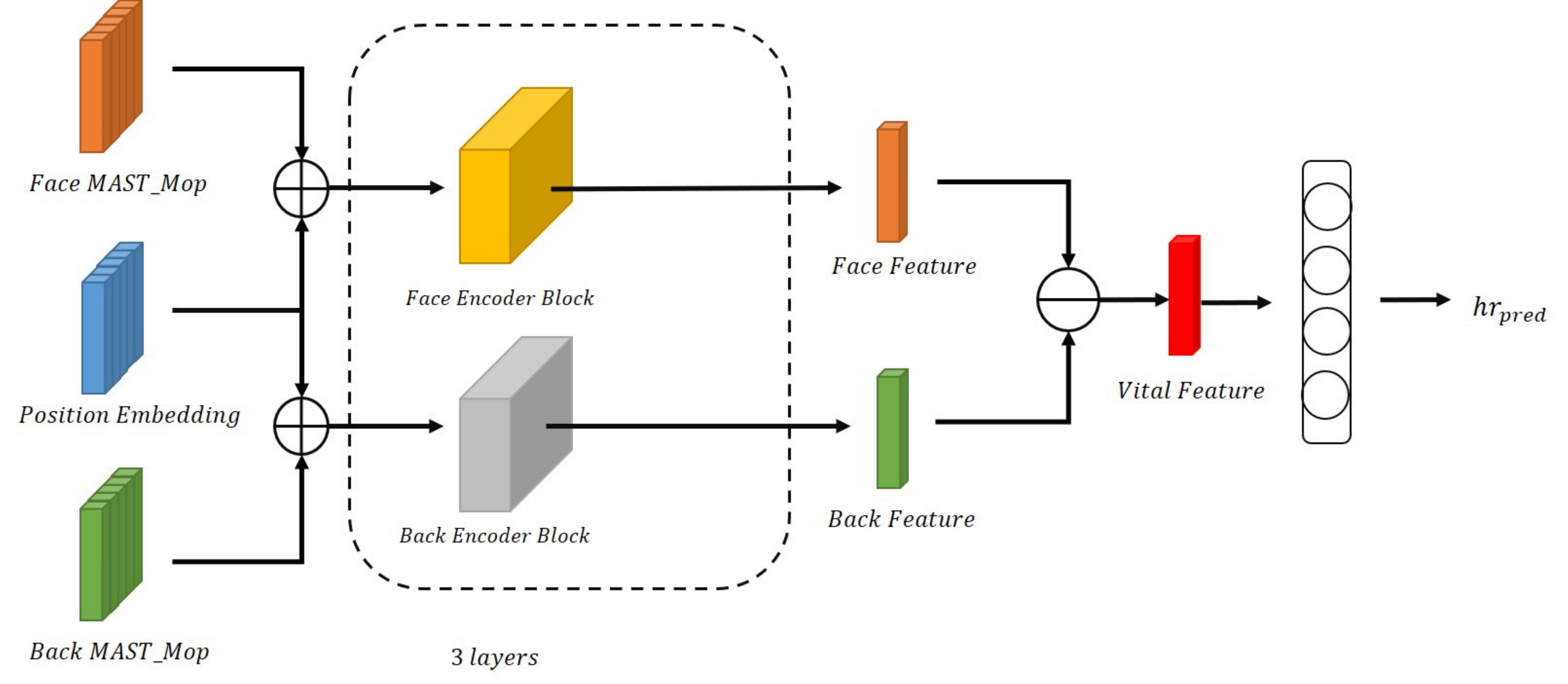}
		\caption{Overview of our framework. We first embed the facial videos into MAST\_Mop, and then feed it to a two-stream Transformer, in which one stream learns the foreground features while the other stream learns the background features. After that, we subtract the background feature from that of foreground, and then utilize a Multilayer Perceptron (MLP) predictor to estimate HR}.
		\label{fig2}
	\end{figure*}
	
	\section{Related Works}
	\subsection{Photoplethysmography}
	Photoplethysmography (PPG) \cite{ksc:pulse} is a widely used contact HR measurement technology. Lots of equipments such as finger-clip HR meter and smart watch are all for PPG applications. Its principle is to calculate HR through measuring the volume change of blood in capillaries on the surface of skin caused by heartbeat. More specifically, as shown in Figure \ref{fig1}, the blood volume in capillaries increases with heart systole  and decreases with diastole. Accroding to the Lambert-Beer law \cite{l:lamnert}, the absorption of illumination by blood varies with its volume while the absorption by other tissues (e.g., dpidermis, dermis) is constant. Therefore, we use constant-intensity light to irradiate the skin surface. When the heart is systolic, the absorption of light increases, so the intensity of reflected light decreases, while opposite during the diastolic time. Accordingly, we can utilize a photosensitive sensor to receive the reflected light. By analyzing the change of intensity, we can get BVP and calculate the heart rate correspondingly.
	\subsection{Remote-photoplethysmography}
	The first study of remote photoplethysmography (rPPG) is reported in \cite{vsn:remote}, which is based on the traditional PPG algorithm. Traditional PPG algorithms require a light source transmitter with constant intensity  and a photosensitive receiver, and both need to be contacted to the skin. In \cite{vsn:remote}, the authors suggested to utilize natural illumination instead of a transmitter and applied a webcamera as the receiver, which is also able to capture change of skin color. Meanwhile, considering the dense capillaries in human faces, rPPG extracts BVP signals from face region instead of from finger as done in traditional PPG algorithms.
	
	However, natural illumination is not constant, many factors such as air humidity and dust will affect the intensity of light. At the same time, motion of head will also have a remarkable impact on the consistency of perceived color signals, so that early studies of rPPG algorithms such as \cite{pmp:no,lczp:remote,vsn:remote} only work well under well-controlled scenes, which is not practical.
	
	In order to improve the accuracy and robustness of rPPG algorithms, many studies have been conducted in recent years, which can be divided into two categories: Traditional hand-crafted methods and learning-based methods. The traditional approaches aim to extract more robust BVP signals using different color spaces \cite{dj:robust,wdsd:pos} or different regions of interest (ROIs) \cite{lczp:remote,tar:sr}. Besides, independent component analysis (ICA) \cite{pmp:no} and principal components analysis (PCA) \cite{lrkn:measuring} are widely used to approach cleaner BVP signals. However, most of these methods are based on some priori assumptions, which may not hold in less-constrained and complicated scenarios, resulting in significant performance degradation in terms of accuracy and stability for HR estimation.
	
	In addition, there are also a number of learning-based rPPG algorithms with the rise of deep learning recently. \cite{cm:deepphys} proposed a CNN model to deal with head motion by using attention mechanism to focus on regions more important so that more stable BVP signals can be extracted. \cite{yplhz:stven} proposed a two-stage model: They utilized an encode-decoder to enhance quality of videos, and then utilized a CNN model to predict HR from the enhanced videos. \cite{nshc:rhythmnet} first proposed a new representation of videos, and then utilized a CNN-RNN model to estimate HR from it. \cite{nyhlsz:cvdnet} proposed a novel framework based on disentangle representation learning so that they can filter out noise more efficiently. \cite{splfz:deephr} proposed a two-stage framework that first learns a discriminative representation of face videos, and then uses an auto-encoder to regress HR value. These methods have made lots of breakthroughs by improving network structure and feature extraction. However, there are also some common defects: Firstly, they extract features only from face regions, which losts background information that can act as noise model to refine foreground information. Secondly, all these approaches utilize CNN-based models, which do not work well in capturing temporal dependencies.
	\subsection{Transformer}
	Transformer is first proposed in \cite{vspujgkp:attention} by Google. The model  stacks self attention layers and fully connection layers for encoder and decoder, without repetition and convolution. It was first utilized in various tasks in natural language processing (NLP), such as the BERT model proposed in  \cite{dclt:bert} that replaces Word2Vec \cite{lm:distributed}. The model utilized Transformer as the backbone, which is able to capture the bidirectional relationship in sentences more thoroughly. Moreover, due to the powerful ability to capture long-term context information, Transformer is also applied to time series processing \cite{ljxzcwy:enhancing}, where the encoder part of Transformer is used to obtain latent features to conduct regression. Hereafter, \cite{zzpzlxz:informer} proposed ProbSparse self-attention mechanism to reduce time complexity and a generative style decoder to acquire long sequence output with only one forward step needed, avoiding cumulative error spreading during the inference phase.

	\begin{figure*}[t]
		\centering
		\includegraphics[width=0.6\textwidth,height=0.2\textheight]{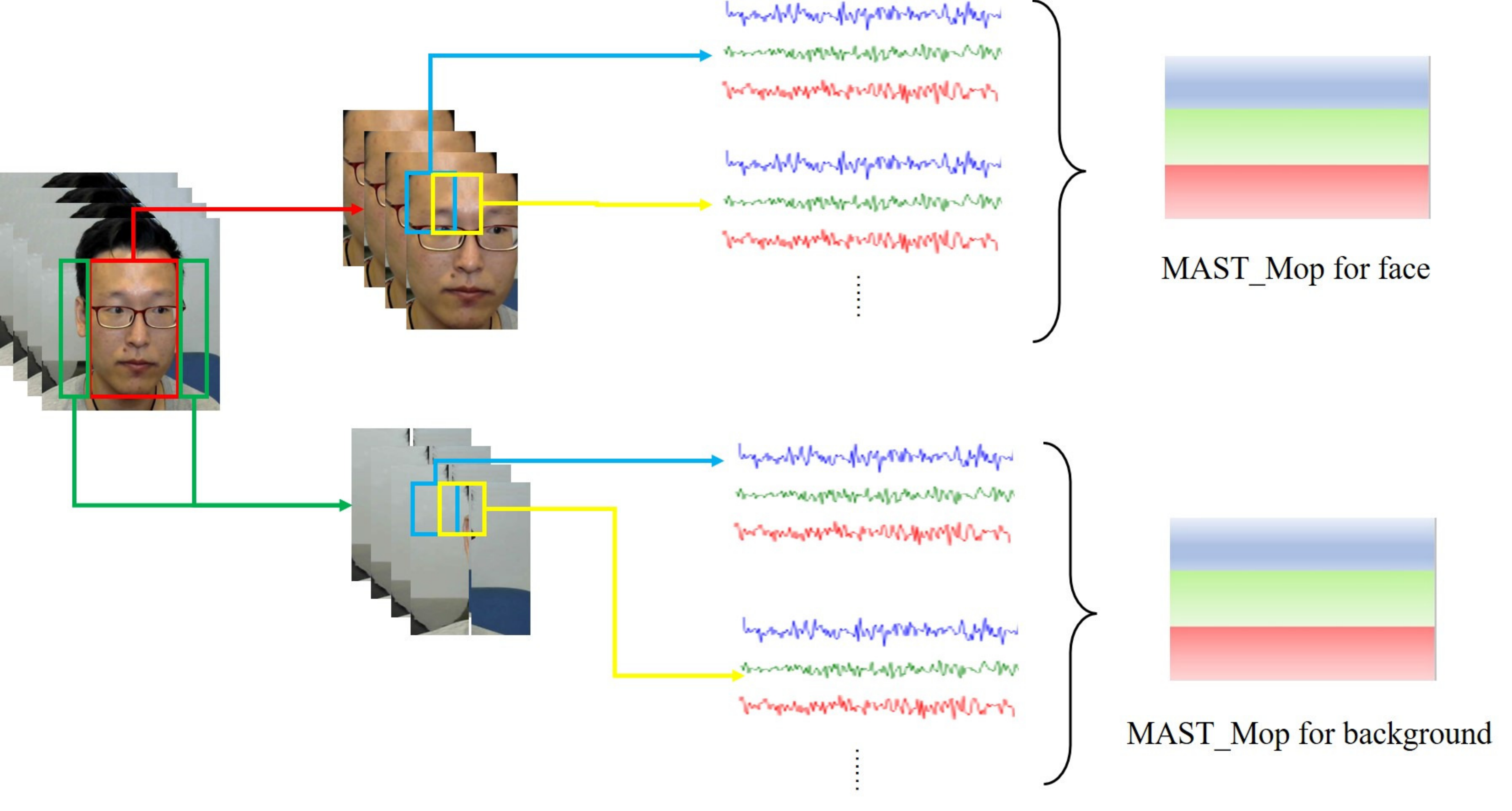}
		\caption{Video embedding: We first get the bounding box of the face (red box) and background (green box) region spanned by the 81 landmarks. Next, we split each of the $T$ cropped frames into $n$ blocks and get $3n$ sequences of length $T$. We repeat the above process $k$ times with different value of $n$, say, $\{n_1, n_2, ..., n_k\}$, and finally get the MAST\_Mop with the size of $3 \sum_{i=1}^{k}n_i \times T$.}
		\label{fig3}
	\end{figure*}

	\section{Method}
	\noindent Figure \ref{fig2} gives an overview of our framework. There are two steps in the framework: First, we embed a facial video into two feature maps referred to as MAST\_Mop that represent the foreground and background information. Second, a two-stream Transformer model is designed to learn a mapping from MAST\_Mop to HR value.
	\subsection{Video embedding}
	\noindent The target of video embedding is to embed a video to a feature map that contains both spatial and temporal information. There are totally two steps in our video embedding approach: First, we crop a particular region of every frame of the video. Second, we compute a spatial map for every frame, and aggregate all spatial maps into a spatiotemporal map MAST\_Mop.

	\subsubsection{Video cropping:}
	For one frame, as shown in Figure \ref{fig3}, we first localize 81 facial landmarks with open source toolkit \emph{Seetaface} \cite{s:seetaface}. Then, a face bounding box is obtained with the width and height to be $w_{face}$ and $h_{face}$, where $w_{face}$ is the horizontal distance between the outer cheek border points and $h_{face}$ is 1.2 times of the vertical distance between chin location and eye eyebrow centre, the same as \cite{nshc:rhythmnet}. This corresponds with foreground but include both vital and non-vital information, which might arise from unstable illumination. Using merely the information from foreground leads to the difficulty in separating vital information from non-vital information if without the reference model of pure non-vital information. Since only non-vital information are contained in background in general, in this study, we let the background serve as the reference model of non-vital information so as to identify its counterpart in the foreground for noise suppressing by comparing foreground to background. Note that such a scheme holds due to the fact that the perturbations on foreground and the surrounding background are in general homogeneous. For this sake, we crop $w_{pad}$ on both left and right side of the face region and concatenate the two cropped regions together to form the background region, which is refer to as $BBox_b$. Specifically, let $$\textbf{\emph{v}} = \{f_i | i = 1, 2, ..., T;  f_i \in \mathbf{R}^{3 \times h \times w}\}$$ indicate a video, where $T$ denote the number of frames, 3 the RGB channels, and $h$, $w$ the height and width of the frame, respectively. We first get the bounding boxes of face regions referred to as $BBox_f$ through the facial landmarks:$$BBox_{f} = \{[x_1^i, y_1^i, x_2^i, y_2^i] | i = 1, 2, ..., T\}$$ where $(x_1, y_1)$ and $(x_2, y_2)$ denotes the upper left point and the lower right point of $BBox_{f}$, respectively. Then, we can get the $BBox_b$ as follows: $$BBox_{b} = \{concat(BBox_{bl}^i, BBox_{br}^i)| i = 1,2,...,T\}$$ $$BBox_{bl} = \{[x_1^i - w_{pad}, y_1^i, x_1^i, y_2^i]| i = 1,2,...,T\}$$ $$BBox_{br} = \{[x_2^i, y_1^i, x_2^i + w_{pad}, y_2^i] | i = 1,2,...,T\}$$ $$w_{pad} = 0.2 \times (x_2^i - x_1^i)$$ Note that this step is not trivial in that it incorporates the background area into the receptive field such that the perturbation caused by illumination instability should be reflected in foreground and background simultaneously and homogeneously due to the adjacency between them.
	\subsubsection{MAST\_Mop for Representing BVP:}
	Our MAST\_Mop is inspired by the MST\_Map introduced in \cite{nshc:rhythmnet}, in which they first apply a sliding-window to partition every frame into $n$ blocks, then concatenate the average pixel value of each block at the corresponding location in a video sequence, and finally concatenate $n$ sequences into a feature map (MST\_Map) to represent the BVP. MST\_Map utilizes the whole face information instead of using only a specific region like \cite{lczp:remote}, making the information more sufficient. Meanwhile, the average pooling over blocks makes it more robust. However, there are still some defects: First, to ensure the number of blocks of each frame identical to guarantee the same feature dimension, they resize the cropped frames to fit a fixed window size, which may introduce new noise arising from image scaling. Second, they apply only one fixed window size when sliding, making the features too monotonous.
	To avoid these, we proposed MAST\_Mop. As shown in Figure \ref{fig3}, the difference from MST are as follows:
	\begin{itemize}[fullwidth,itemindent=2em]
		
		\item[(1)] Since resize operation is noise prone, instead of performing it to normalize the cropped regions of different sizes, we apply a fixed number of sliding windows to cover a cropped region such that the size of the sliding windows is adapted to the size of the cropped region, which avoids the noise-inducing resize operation. In detail, the size of the sliding window is subject to both the number of sliding windows and the size of the region cropped from the $i$th frame, defined as: $$ws^i=(2\frac{w_c^i}{\sqrt{n} + 1}, 2\frac{h_c^i}{\sqrt{n} + 1})$$ with the overlap between two sliding windows to be:$$step^i = (\frac{w_{c}^i}{\sqrt{n} + 1}, \frac{h_{c}^i}{\sqrt{n} + 1})$$ where $h_c^i$ and $w_c^i$ denote the height and width of the $i$th $BBox_f$ or $BBox_b$.
		\item[(2)]  To further grant the robustness of feature extraction, we apply a couple of numbers of sliding windows to incorporate different resolutions, that is,$$n = n_1, n_2, ..., n_k$$For each $n_i$, we calculate the corresponding feature map:$$feat = \{feat_i \in \mathbf{R}^{3n_i \times T} | i=1,2, ..., k\}$$  Then, we concatenate all the maps to obtain the multiscale feature map to promise robustness against the varying size of ROI across frames in various scenarios: $$Feat \in \mathbf{R^{3\sum_{i=1}^kn_i \times T}}$$
		
	\end{itemize}
%	\begin{table*}[t]
%		\centering
%		\begin{tabular}{c c c c | }
%			\hline
%			Methods  & RMSE & Std & $r$ \\
%			\hline\hline
%			\cite{pmp:no}       & 13.6  & 24.3  & 0.08 \\
%			\cite{dj:robust}    & 22.36 & 13.67 & 0.21 \\
%			\cite{lczp:remote}  & 7.62  & 6.88  & 0.81 \\
%			\cite{tar:sr}       & 6.23  & 5.81  & 0.83 \\
%			\cite{yplhz:stven}  & 5.93  & 5.57  & 0.88 \\
%			\cite{ylnsz:autohr} & 5.10  & $\mathbf{4.73}$  & 0.86\\
%			\hline\hline
%			Ours  &$\mathbf{4.83}$ & 4.77 & $\mathbf{0.92}$\\
%			\hline
%			
%		\end{tabular}
%		\caption{Comparison of our method with state-of-the-art methods for HR
%			estimation on MAHNOB-HCI dataset. Values in bold show the best result}
%		\label{table1}
%	\end{table*}
%	\begin{table*}[t]
%		\centering
%		\begin{tabular}{c c c c c}
%			\hline
%			Methods & MAE & RMSE & Std & $r$ \\
%			\hline\hline
%			\cite{dj:robust}       & 11.4 & 16.9 & 15.1 & 0.28 \\
%			\cite{tar:sr}          & 11.5 & 17.2 & 15.3 & 0.30 \\
%			\cite{cm:deepphys}     & 11.0 & 13.8 & 13.6 & 0.11 \\
%			\cite{nshc:rhythmnet}  & 5.30 & 8.14 & 8.11 & 0.76 \\
%			\cite{nyhlsz:cvdnet}   & 5.02 & 7.97 & 7.92 & $\mathbf{0.79}$ \\
%			\hline\hline
%			Ours  & $\mathbf{4.94}$ & $\mathbf{7.31}$ & $\mathbf{7.30}$ & $\mathbf{0.79}$\\
%			\hline
%			
%		\end{tabular}
%		\caption{Comparison of our method with state-of-the-art methods for HR
%			estimation on VIPL dataset. Values in bold show the best result}
%		\label{table2}
%	\end{table*}

	\begin{table*}[t]
		\centering
		\begin{tabular}{c | c c c | c c c c}
		\hline
		&&MAHNOB-HCI&&&VIPL-HR&&\\
		\hline\hline
		                     & RMSE  & std   & r     & MAE & RMSE &std &r \\
		\hline
		\cite{pmp:no}        & 13.6  & 24.3  & 0.08  &  -   &-     &-     &-\\
		\cite{dj:robust}     & 22.36 & 13.67 & 0.21  & 11.4 & 16.9 & 15.1 & 0.28\\
		\cite{lczp:remote}   & 7.62  & 6.88  & 0.81  &  -   &-     &-     &-\\
		\cite{tar:sr}        & 6.23  & 5.81  & 0.83  & 11.5 & 17.2 & 15.3 & 0.30\\
		\cite{cm:deepphys}   & -     & -     &  -    & 11.0 & 13.8 & 13.6 & 0.11\\
		\cite{yplhz:stven}   & 5.93  & 5.57  & 0.88  &  -   &-     &-     &-\\
		\cite{ylnsz:autohr}  & 5.10  & $\mathbf{4.73}$  & 0.86 &  -   &-     &-     &-\\
		\cite{nyhlsz:cvdnet} & 5.23 & 5.10 & 7.92  & 5.02 & 7.97 & 7.92 & $\mathbf{0.79}$\\
		Ours &$\mathbf{4.88}$ & 4.80 & $\mathbf{0.92}$& $\mathbf{4.94}$ & $\mathbf{7.42}$ & $\mathbf{7.44}$ & $\mathbf{0.79}$\\
		\hline
		\end{tabular}
		\caption{Comparison of our method with state-of-the-art methods for HR
			estimation. Values in bold show the best result}
		\label{table1}
	\end{table*}

	\subsection{Two-stream Transformer}
	As concluded in \cite{lczp:remote}, the features extracted from foreground can be regarded as superposition of non-vital information (such as illumination changes) over vital information (BVP), while the features extracted from background only contains non-vital information. This motivates us to let the system be aware of the non-vital information disturbing the foreground by referring to the background. To approach this goal, we proposed a  network being composed of two independent streams. As shown in Figure \ref{fig2}, the two streams both take MAST\_Mop as their input. The fore-stream ($\mathbf{FS}$) aims to learn foreground features, while the back-stream ($\mathbf{BS}$)  background features. After that, we subtract the background features from foreground features so as to get stable and cleaner BVP features. Since background and foreground are adjacent to each other, the non-vital information such as unstable illumination should be imposed on them simultaneously and homogeneously, this makes the subtraction between the two streams work in a way like adaptive noise cancellation. Finally, we utilize a multilayer perceptron ($\mathbf{MLP}$) as the predictor to estimate HR from the BVP features, trained under the supervision of the ground-truth HR with $L1$ loss function.
	
	It is also worth noting that to deal with the long-range temporal dependencies in BVP signals, we utilize the encoder proposed in Informer \cite{zzpzlxz:informer} to extract features from MAST\_Mop rather than the primordial Transformer encoder \cite{vspujgkp:attention}. After the video embedding, we have got an feature map $Feat \in \mathbf{R^{3\sum_{i=1}^kn_i \times T}}$. However, considering the problem that the HR could change remarkably if the video is too long while the granularity of HR is too coarse to be precisely captured if the video is too short, we cut the whole video into small clips of fixed length and obtain the corresponding features $Feat_c \in \mathbf{R^{3\sum_{i=1}^kn_i \times t}}$, where $t$ is the frame number of one video clip. Then, we feed $Feat_c$ as the input to the network, and then obtain the corresponding heart rate under the supervision: $$Loss = \sum |HR_{gt} - \mathbf{MLP}(\mathbf{FS}(Feat_c) - \mathbf{BS}(Feat_c))|$$For long term estimating, we just perform the estimation for each clip separately, and then compute the average value of all the clips.
	
	\section{Experiments}
	\noindent In this section, we provide evaluations of the proposed framework on two widely used datasets: MAHNOB-HCI \cite{slpp:hci} and VIPL-HR \cite{nhsc:vipl}.
	
	\subsection{Datasets}
	$\mathbf{MAHNOB-HCI}$ \cite{slpp:hci} is a multimodal dataset with videos of subjects participating in two experiments: Emotion elicitation and implicit tagging. It contains 27 subjects (12 males and 15 females) in total, and all 527 videos are in 780x580 resolution along with the corresponding ECG signals.

	\noindent $\mathbf{VIPL-HR}$ \cite{nhsc:vipl} is a large public dataset, whose target is to promote researches on remote HR estimation under less-constrained scenarios, such as with head movement, illumination variation, and acquisition device diversity. VIPL-HR contains 2371 RGB facial videos from 107 subjects, which are recorded with three different devices (webcam, RealSense, and smartphone) under varying illuminations and pose variations. Meanwhile, it offers the ground truth HR records.
	\begin{figure}[h]
		\centering
		\includegraphics[width=0.9\columnwidth]{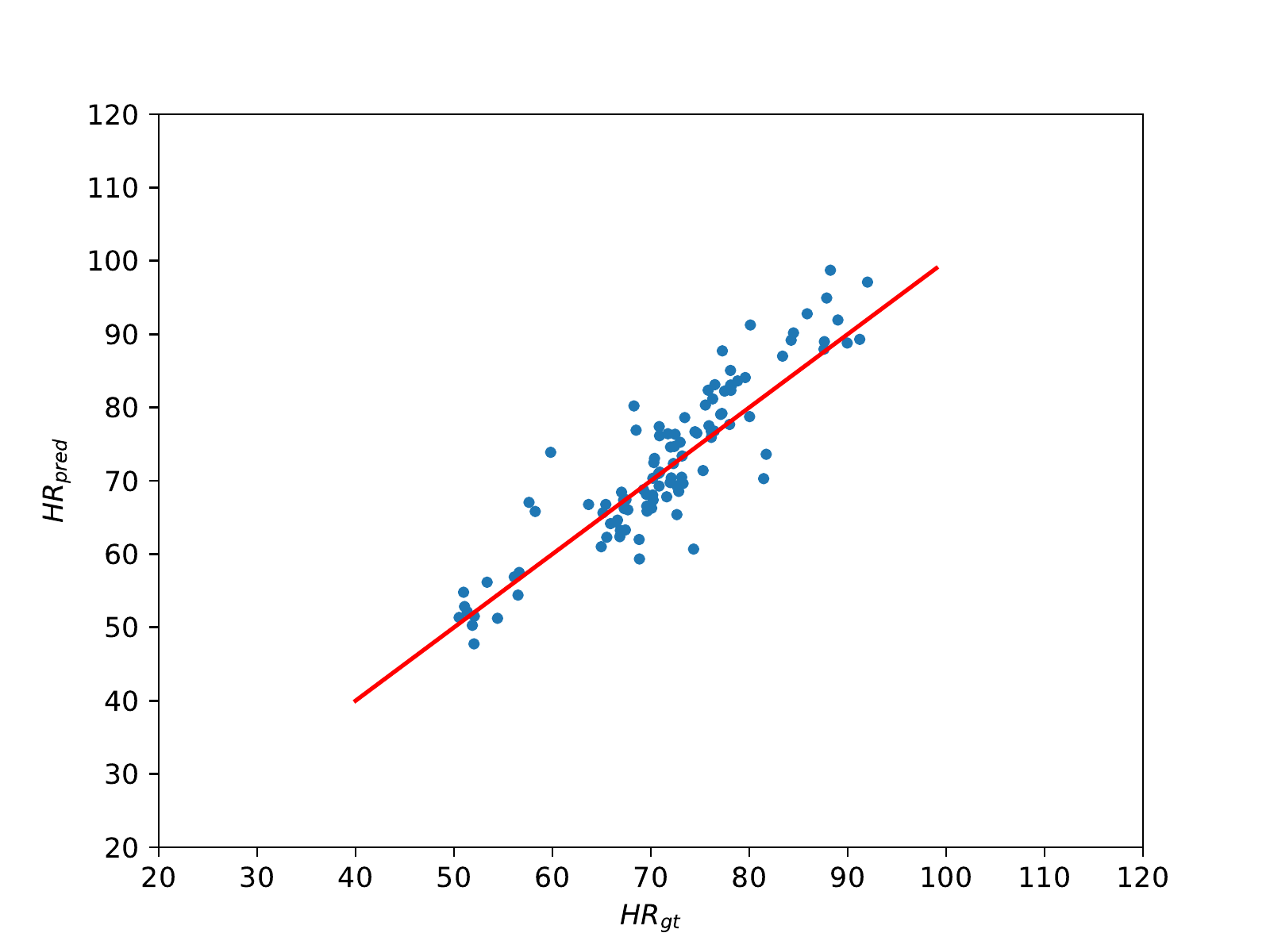}
		\caption{Scatter plot camparing the ground truth $HR_{gt}$ and the predicted  $HR_{pred}$ on MAHNOB-HCI}.
		\label{fig4}
	\end{figure}
	\subsection{Experimental setting}
	We implement our framework with PyTorch \cite{pgccydldal:pytorch} on a GeForce GTX2080Ti GPU. We set $n = \{25, 81, 169\}$ for the sliding window and both of the fore-stream ($\mathbf{FS}$) and back-stream ($\mathbf{BS}$) are composed of 3 standard Informer encoders \cite{zzpzlxz:informer}. For MAHNOB-HCI \cite{slpp:hci}, we compute the ground truth HR of the second channel (EXG2) through the ECG waveforms provided in the data. For training, every video is cut into 10-second clips with a moving forward step of 0.5 second. For testing, following early works such as \cite{lczp:remote} and \cite{nshc:rhythmnet}, we use a 30-second clip (from 10 second to 40 second) to evaluate our model. As for VIPL \cite{nhsc:vipl}, we do the same as MAHNOB-HCI \cite{slpp:hci} when training and use a 20-second (from 0 to 20 second) cilp for testing as the durations of many videos are shorter than 30 seconds in the dataset. We use adam optimizer \cite{kb:14} to train our framework for 80 epochs with the learning rate set to 0.0001 for MAHNOB-HCI and 120 epochs with the learning rate set to 0.00005 for VIPL-HR.
	
	We finally compare our framework with a couple of methods on multiple metrics such as mean absolute error (MAE), root mean square erros (RMSE), mean error (Mean) and standard deviation (Std), and Pearson correlation coefficients ($r$).
	
	\subsection{Results}
	
	\subsubsection{Results on MAHNOB-HCI:} We use 422 videos for training and 105 videos for test. The performance of our method compared with the other state-of-art methods are shown in Table \ref {table1}. Also, we plot the camparison between the ground truth $HR_{gt}$ and the predicted $HR_{pred}$ in Figure \ref{fig4}. Note that \cite{nshc:rhythmnet,yplhz:stven,splfz:deephr} are all learning-based methods. As can be seen, our method obtains substantially better results than these baseline methods with a considerable progress. From Table \ref{table1} and Figure \ref{fig4}, we can see that the outcome of our framework in estimating HR is more accurate and more stable, no large deviation in the prediction results.

	\subsubsection{Results on VIPL:} 
\begin{figure}[h]
		\centering
		\includegraphics[width=0.9\columnwidth]{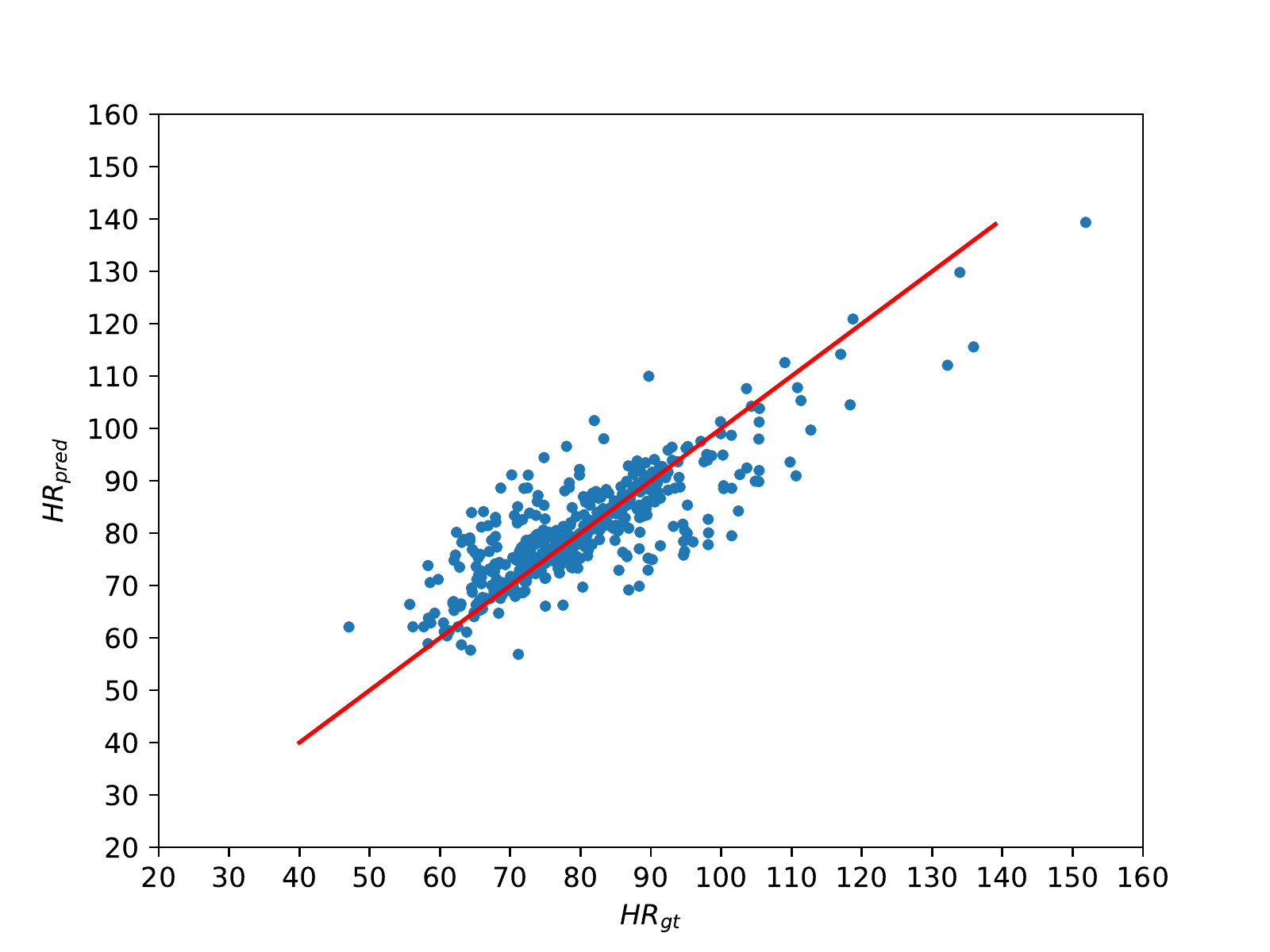}
		\caption{Scatter plot camparing the ground truth $HR_{gt}$ and the predicted  $HR_{pred}$ on VIPL-HR}.
		\label{fig5}
	\end{figure}

We use 1910 videos for training and 468 videos for test. The performance of our method compared with those of the other state-of-art methods are shown in Table \ref{table1}. Also, we plot the camparison between the ground truth $HR_{gt}$ and the predicted $HR_{pred}$ in Figure \ref{fig5}. Note that all methods referred to in Table \ref{table1} are learning-based except \cite{dj:robust} and \cite{tar:sr}. From the results, we can see that our method has achieved a much higher accuracy with $MAE$=4.94 and $RMSE$=7.42. Besides, the lowest $Std$=7.30 shows the better stability of our method.
	
	\subsection{Ablation experiments}
	In order to validate the effectiveness of the two-stream network, we use MAST\_Mop as input to train a one-stream network containing only foreground stream ($\mathbf{FS}$) so as to check whether the proposed method really benefits from the two-stream based adaptive noise suppressing. Furthermore, to validate the contribution of the MAST\_Mop, we use STMap \cite{nshc:rhythmnet} as input instead of MAST\_Mop to train this one-stream network so as to check whether the performance varies with feature. Feature comparison on the two-stream framework is also conducted. The results are shown in Table \ref{table3}, from which we can see that the performance on the two datasets RMSE is increased by 0.54 and 0.68 by applying the two stream structure and MAST\_Mop.
	\begin{table}[t]
		\centering
		\begin{tabular}{p{80pt} | c | c c}
			\hline
			&\multicolumn{1}{c|}{MAHNOB-HCI}&\multicolumn{2}{c}{VIPL-HR}\\
			\hline\hline
			& RMSE & MAE & RMSE \\
			\hline
			$\mathbf{FS}$ + STMap  & 5.42 & 5.12 & 8.10 \\
			$\mathbf{FS}$ + $\mathbf{BS}$ + STMap  & 5.18 & 5.62 & 7.71 \\
			$\mathbf{FS}$ + MAST\_Mop  & 5.23 & 5.10 & 7.92 \\
			TransPPG & $\mathbf{4.88}$ & $\mathbf{4.94}$ & $\mathbf{7.42}$ \\
			\hline
			
		\end{tabular}
		\caption{Ablation experiments.}
		\label{table3}
	\end{table}
	
	\section{Conclusion}
	In this paper, we proposed a novel framework for accurate and stable HR estimation under less-constrained scenarios, which achieves remarkably better results under multiple metrics. Our framework advances the literature with two innovations: Video Embedding and Two-stream Transformer. Video embedding aims to map the video to a feature map that represent the vital and non-vital information with multi-resolution window size but avoid the noise-prone resize operation at the same time. Two-stream Transformer is used to estimate HR with the vital and non-vital information so as to reduce the interference by incorporating the two channels as the reference profile to each other. The reason that we do not use CNN-based models is that Transformer-based models are better at capturing long-term context in the extracted feature maps. As far as we konw, it is the first work with Transformer as backbone to capture the temporal dependencies in rPPGs and apply the two stream scheme to figure out the interference from background as mirror of the corresponding perturbation on foreground signals. We evaluate our framework on two widely used datasets and the results shows that our framework outperforms the state-of-art methods on the two widely used benchmarks.
	%\appendix
	%\label{sec:reference}
	%\nobibliography*
	% Use \bibliography{yourbibfile} instead or the References section will not appear in your paper
	%\bibliography{ref}
	\bibliographystyle{named}
	\bibliography{refe}
\end{document}